\pdfoutput=1
\documentclass[a4paper,12pt]{article}
\usepackage{lmodern}
\usepackage[left=2.5cm,right=2.5cm,top=2.5cm,bottom=3cm]{geometry}
\usepackage{textcomp}
\usepackage[compact,center]{titlesec}

\def\PitmanYor{Pitman--\kern-0.13em{}Yor}

\usepackage[authoryear,square]{natbib}
\usepackage{cleveref,setspace}
\usepackage[lofdepth,lotdepth,caption=false]{subfig} 

\usepackage{hyperref}
\hypersetup{
colorlinks=true,
 citecolor=blue,
 linkcolor=red,
 urlcolor=magenta}

\usepackage{amsmath,amsthm,amsfonts,mathtools,bbm,graphicx,ifpdf,booktabs,enumerate}
\newcommand{\nwc}{\newcommand}
\newcommand{\rnwc}{\renewcommand}
\newcommand{\nwt}{\newtheorem}
\newcommand{\bC}{\hbox{{\rm I}
                   \kern-.8em\hbox{{\rm C}}}} 
\nwc{\ba}{\begin{array}}
\nwc{\be}{\begin{equation}}
\nwc{\beq}{\begin{eqnarray}}
\nwc{\beqn}{\begin{eqnarray*}}
\nwc{\beqast}{\begin{eqnarray*}}
\nwc{\bm}{\boldmath}

\nwc{\m}{\mbox}
\nwc{\ubm}{\unboldmath}

\nwc{\bma}{\m{\bm $a$ \ubm}}
\nwc{\bmma}{\m{\bm $a$\ubm}}
\nwc{\bmb}{\m{\bm $b$ \ubm}}
\nwc{\bmmb}{\m{\bm $b$\ubm}}
\nwc{\bmc}{\m{\bm $c$ \ubm}}
\nwc{\bmmc}{\m{\bm $c$\ubm}}
\nwc{\bmd}{\m{\bm $d$ \ubm}}
\nwc{\bmmd}{\m{\bm $d$\ubm}}
\nwc{\bme}{\m{\bm $e$ \ubm}}
\nwc{\bmme}{\m{\bm $e$\ubm}}
\nwc{\bmf}{\m{\bm $f$ \ubm}}
\nwc{\bmmf}{\m{\bm $f$\ubm}}
\nwc{\bmg}{\m{\bm $g$ \ubm}}
\nwc{\bmmg}{\m{\bm $g$\ubm}}
\nwc{\bmh}{\m{\bm $h$ \ubm}}
\nwc{\bmmh}{\m{\bm $h$\ubm}}
\nwc{\bmi}{\m{\bm $i$ \ubm}}
\nwc{\bmmi}{\m{\bm $i$\ubm}}
\nwc{\bmj}{\m{\bm $j$ \ubm}}
\nwc{\bmmj}{\m{\bm $j$\ubm}}
\nwc{\bmk}{\m{\bm $k$ \ubm}}
\nwc{\bmmk}{\m{\bm $k$\ubm}}
\nwc{\bml}{\m{\bm $l$ \ubm}}
\nwc{\bmml}{\m{\bm $l$\ubm}}
\nwc{\bmm}{\m{\bm $m$ \ubm}}
\nwc{\bmmm}{\m{\bm $m$\ubm}}
\nwc{\bmn}{\m{\bm $n$ \ubm}}
\nwc{\bmmn}{\m{\bm $n$\ubm}}
\nwc{\bmo}{\m{\bm $o$ \ubm}}
\nwc{\bmmo}{\m{\bm $o$\ubm}}
\nwc{\bmp}{\m{\bm $p$ \ubm}}
\nwc{\bmmp}{\m{\bm $p$\ubm}}
\nwc{\bmq}{\m{\bm $q$ \ubm}}
\nwc{\bmmq}{\m{\bm $q$\ubm}}
\nwc{\bmr}{\m{\bm $r$ \ubm}}
\nwc{\bmmr}{\m{\bm $r$\ubm}}
\nwc{\bms}{\m{\bm $s$ \ubm}}
\nwc{\bmms}{\m{\bm $s$\ubm}}
\nwc{\bmt}{\m{\bm $t$ \ubm}}
\nwc{\bmmt}{\m{\bm $t$\ubm}}
\nwc{\bmu}{\m{\bm $u$ \ubm}}
\nwc{\bmv}{\m{\bm $v$ \ubm}}
\nwc{\bmmv}{\m{\bm $v$\ubm}}
\nwc{\bmw}{\m{\bm $w$ \ubm}}
\nwc{\bmmw}{\m{\bm $w$\ubm}}
\nwc{\bmx}{\m{\bm $x$ \ubm}}
\nwc{\bmmx}{\m{\bm $x$\ubm}}
\nwc{\bmy}{\m{\bm $y$ \ubm}}
\nwc{\bmmy}{\m{\bm $y$\ubm}}
\nwc{\bmz}{\m{\bm $z$ \ubm}}
\nwc{\bmmz}{\m{\bm $z$\ubm}}

\nwc{\bmA}{\m{\bm $A$\ubm}}
\nwc{\bmB}{\m{\bm $B$\ubm}}
\nwc{\bmD}{\m{\bm $D$\ubm}}
\nwc{\bmF}{\m{\bm $F$\ubm}}
\nwc{\bmI}{\m{\bm $I$\ubm}}
\nwc{\bmJ}{\m{\bm $J$\ubm}}
\nwc{\bmM}{\m{\bm $M$\ubm}}
\nwc{\bmP}{\m{\bm $P$\ubm}}
\nwc{\bmR}{\m{\bm $R$\ubm}}
\nwc{\bmU}{\m{\bm $U$\ubm}}
\nwc{\bmX}{\m{\bm $X$\ubm}}

\nwc{\bmalpha}{\m{\bm  $\alpha$  \ubm}}
\nwc{\bmeta}{\m{\bm    $\eta$    \ubm}}
\nwc{\bmgamma}{\m{\bm  $\gamma$  \ubm}}
\nwc{\bmkappa}{\m{\bm  $\kappa$  \ubm}}
\nwc{\bmlambda}{\m{\bm $\lambda$ \ubm}}
\nwc{\bmmu}{\m{\bm $\mu$\ubm}}
\nwc{\bmnabla}{\m{\bm  $\nabla$  \ubm}}
\nwc{\bmnu}{\m{\bm     $\nu$     \ubm}}
\nwc{\bmomega}{\m{\bm  $\omega$  \ubm}}
\nwc{\bmphi}{\m{\bm    $\phi$    \ubm}}
\nwc{\bmpsi}{\m{\bm    $\psi$    \ubm}}
\nwc{\bmsigma}{\m{\bm  $\sigma$  \ubm}}
\nwc{\bmtau}{\m{\bm    $\tau$    \ubm}}
\nwc{\bmtheta}{\m{\bm  $\theta$  \ubm}}
\nwc{\bmzeta}{\m{\bm  $\zeta$  \ubm}}
 \nwc{\bmxi}{\m{\bm    $\xi$     \ubm}}

\nwc{\bmGamma}{\m{\bm $\Gamma$\ubm}}
\nwc{\bmLambda}{\m{\bm $\Lambda$\ubm}}
\nwc{\bmPhi}{\m{\bm $\Phi$\ubm}}

\nwc{\ca}{{\cal A}}
\nwc{\cao}{{\cal A}^{-1}}
\nwc{\cb}{{\cal B}}
\nwc{\cc}{{\cal C}}
\nwc{\cd}{{\cal D}}
\nwc{\ce}{{\cal E}}
\nwc{\cf}{{\cal F}}
\nwc{\cg}{{\cal G}}
\nwc{\ch}{{\cal H}}
\nwc{\ci}{{\cal I}}
\nwc{\cj}{{\cal J}}
\nwc{\ck}{{\cal K}}
\nwc{\cl}{{\cal L}}
\nwc{\clu}{{\cal L}{\cal U}}
\nwc{\cm}{{\cal M}}
\nwc{\cn}{{\cal N}}
\nwc{\co}{{\cal O}}
\nwc{\cp}{{\cal P}}
\nwc{\cq}{{\cal Q}}
\nwc{\calr}{{\cal R}}
\nwc{\cs}{{\cal S}}
\nwc{\ct}{{\cal T}}
\nwc{\cu}{{\cal U}}
\nwc{\cv}{{\cal V}}
\nwc{\cw}{{\cal W}}
\nwc{\cx}{{\cal X}}
\nwc{\cy}{{\cal Y}}
\nwc{\cz}{{\cal Z}}

\nwc{\ea}{\end{array}}
\nwc{\ee}{\end{equation}}
\nwc{\eeq}{\end{eqnarray}}
\nwc{\eeqn}{\end{eqnarray*}}
\nwc{\eeqast}{\end{eqnarray*}}

\nwc{\half}{\frac{1}{2}}

\nwc{\noi}{\noindent}
\nwc{\non}{\nonumber}
\nwc{\onehalf}{\frac{1}{2}}
\nwc{\onethird}{\frac{1}{3}}

\nwc{\p}{\partial}

\nwc{\uP}{{\em \bf Proof: }}
\nwc{\uT}{\underline{Theorem:}}

\nwt{thm}{Theorem}
\nwt{lem}[thm]{Lemma}
\nwt{prop}[thm]{Proposition}
\nwt{cor}[thm]{Corollary}

\theoremstyle{definition}
\ifdefined\definition  \else  \nwt{definition}[thm]{Definition}  \fi
\nwt{conj}{Conjecture}[section]
\nwt{exmp}{Example}[section]
\nwt{rem}[thm]{Remark}

\theoremstyle{remark}
\nwt{note}{Note}
\nwt{case}{Case}

\nwc{\mcal}{\mathcal}
\rnwc{\vec}[1]{\boldsymbol{\mathbf{#1}}}

\nwc{\Ints}{\mathbb{Z}}
\nwc{\NonNegInts}{\mathbb{Z}_+}
\nwc{\Nats}{\mathbb{N}}
\nwc{\Rationals}{\mathbb{Q}}
\nwc{\Reals}{\mathbb{R}}
\nwc{\kernel}{\kappa}
\nwc{\kernelmatrix}{K}
\nwc{\logistic}{\sigma}

\nwc{\Normal}{\mbox{Normal}}
\nwc{\DP}{\mbox{DP}}

\def\BetaDist{\mbox{\rm Beta}}

\def\GammaDist{\mbox{\rm Gamma}}

\nwc{\as}{\textrm{a.s.}}
\nwc{\defas}{:=}

\nwc{\dist}{\ \sim\ }
\nwc{\distiid}{\stackrel{\mathrm{iid}}{\sim}}
\nwc{\disteq}{\stackrel{\mathrm{d}}{=}}

\def\ie{i.e.,\ }
\def\eg{e.g.,\ }
\def\iid{i.i.d.\ }

\numberwithin{equation}{section}
\numberwithin{thm}{section}

\def\[#1\]{\begin{align}#1\end{align}}

\ifpdf
    \graphicspath{{Figures/PNG/}{Figures/PDF/}{Figures/}}
\else
    \graphicspath{{Figures/EPS/}{Figures/}}
\fi

\def\Normal{\mbox{\rm Normal}}
\def\GammaDist{\mbox{\rm Gamma}}

\begin{document}
\title{Dirichlet Fragmentation Processes: \\ A Useful Variant of Fragmentation Processes for Modelling Hierarchical Data}

\author{
Hong Ge \\
University of Cambridge\\
\texttt{hg344@cam.ac.uk} \\
\and
Yarin Gal\\
University of Cambridge \\
\texttt{yg279@cam.ac.uk}
\and
Zoubin Ghahramani \\
University of Cambridge\\
\texttt{zg201@cam.ac.uk} \\
}
\maketitle
\begin{abstract}
Tree structures are ubiquitous in data across many domains, and many datasets are naturally modelled by unobserved tree structures. In this paper, first we review the theory of random fragmentation processes \citep{Bertoin2006Random}, and a number of existing methods for modelling trees, including the popular nested Chinese restaurant process (nCRP). Then we define a general class of probability distributions over trees: the \emph{Dirichlet fragmentation process} (DFP) through a novel combination of the theory of Dirichlet processes and random fragmentation processes. 
This DFP presents a stick-breaking construction, and relates to the nCRP in the same way the Dirichlet process relates to the Chinese restaurant process.
Furthermore, we develop a novel hierarchical mixture model with the DFP, and empirically compare the new model to similar models in machine learning.
Experiments show the DFP mixture model to be convincingly better than existing state-of-the-art approaches for hierarchical clustering and density modelling.
\end{abstract}

\noindent The process of random fragmentation is common to many areas, such as the degradation of large polymer chains in chemistry, or the evolution of phylogenetic trees in biology. An elegant mathematical tool for describing such phenomena is the \emph{fragmentation process} (FP) \citep{Bertoin2006Random}. As a concrete example of a FP, consider a stick of unit length. At every time point, the stick breaks into two smaller pieces. Then, each of the resulting smaller sticks independently repeats the procedure, and the process continues ad infinitum. This process can be described with the FP framework, and generalised to arbitrary distributions over the splits of the stick, breaking times, and number of splits.

The process of fragmentation can be interpreted as inducing a tree structure. 
In the probability theory community, \citet{aldous1991continuum} has worked on binary fragmentation trees and used a symmetric beta distribution as the fragmentation operator for binary trees.
\citet{mccullagh2008gibbs} has worked on the theoretical aspect of \citet{Bertoin2006Random}'s relation to tree priors, studying both binary and multifurcating trees. 
\citet{teh2011modelling} have recently began studying the relation between fragmentation and coagulation processes, and relating these to practical applications in machine learning. 
Apart from the last work, the literature has mostly concentrated on theoretical aspects of the FP, and pragmatic aspects of the process have been largely overlooked.

The rest of this paper is organised as follows. Section \ref{preliminaries} briefly reviews the result of fragmentation processes (FP) as introduced in \citet{Bertoin2006Random}, and the nested Chinese restaurant process \citep{blei2010nested}.
This lays the way for a general probabilistic framework for modelling trees.  
In Section \ref{sec-dfp}, we derive a useful variant of fragmentation processes -- the \emph{Dirichlet fragmentation process} (DFP) -- through a combination of the theory of Dirichlet processes and fragmentation process. A notable property of the DFP is that it relates to the nCRP in the same way the Dirichlet process relates to the Chinese restaurant process, that is the DFP forms the underlying de Finetti measure of the nCRP. Inspired by this property, in Section \ref{sec-model} we develop a hierarchical infinite mixture model with the DFP prior as its mixing distribution, in the same spirit as using the Dirichlet process prior as the mixing distribution for an infinite mixture model. Furthermore, in Section \ref{sec-inference} we describe an associated effective yet simple sampling procedure for the DFP mixture model. Finally, in Section \ref{sec-results} we assess the model with a set of experiments for density estimation and hierarchical clustering, demonstrating an improvement on existing state-of-the-art approaches. 

\section{Preliminaries}
\label{preliminaries}
We begin by briefly reviewing the fragmentation process and nested Chinese restaurant process upon which our new model is based. The relation between the two will become clear in the next sections.

Throughout this paper, we will use finite-length sequences of natural numbers as our index set on the nodes in a tree, i.e. we let $\vec{\omega}=(\omega_1, \omega_2, \ldots, \omega_L)$ denote a length-$L$ sequence of positive integers, $\omega_l \in \Nats$. We denote the zero-length string as $\vec{\omega}=\Delta$ and use $|\vec{\omega}|$ to indicate the length of sequence $\vec{\omega}$. 
When viewing these strings as node indices in a tree, $(\vec{\omega}\omega_i\text{: }\omega_i\in \Nats)$ are the children of $\vec{\omega}$, and $\Delta \preceq \vec{\omega}'\prec\vec{\omega}$ are the ancestors of $\vec{\omega}$, and $\Delta$ is the root of the tree.

\begin{figure*}[t]
\centering
\includegraphics[width=0.52\textwidth]{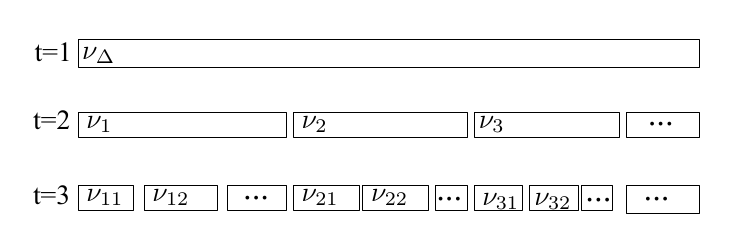}
\includegraphics[width=0.225\textwidth]{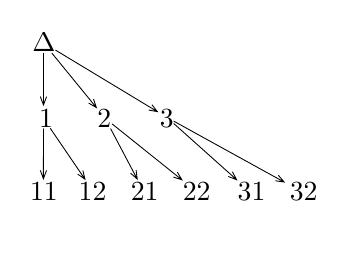}
\includegraphics[width=0.225\textwidth]{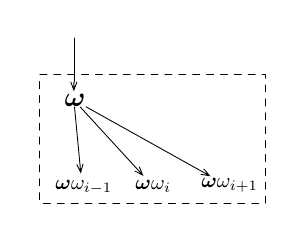}
\vspace{-0.4cm}
\caption{Recursive Stick Breaking. The plot on the left shows an example of recursive breaking; At the first level, the unit-size stick breaks into infinitely many sub-sticks. The first 3 sticks are illustrated and the remaining sticks are represented by dots. Then at the second level, a similar stick breaking process is applied to each sub-stick. This recursive stick-breaking process repeats until a pre-determined maximum depth is reached. The plot in the middle shows the resulting tree structure by discarding stick sizes. The plot on the right shows the sequence indexing scheme.}
\label{fig-re-stick-breaking}
\end{figure*}

\subsection{Fragmentation Processes}
\label{sec-frag-intro}
To give a more concrete description fragmentation processes, first recall the stick-breaking example of fragmentation processes.
We use $\vec{\pi}(t)$ to denote the set of sub-sticks present at each time $t\in \Reals^{+}$ (the set of non-negative real numbers), that is $\vec{\pi}(t) = (\pi_n(t))_{n\in\Nats}$ where the subscript $n$ indexes resulting sub-sticks. Then the stick-breaking process $\Pi=(\vec{\pi}(t))_{t\in \Reals^+}$ is an example of a (mass) fragmentation process.
Motivated by this informal description, we define a fragmentation operator on sequences of real numbers in the general setting, and then give a formal definition of a (mass) fragmentation process over some space $\mcal{S}$. We do this by adapting the formulation in \citep[p. 119]{Bertoin2006Random}.

First consider the space $\mcal{S}$ of non-increasing non-negative sequences that sum to one given by
\(
\mcal{S} \defas
\left\lbrace \vec{\pi} = (\pi_i)_{i\in \Nats} | \pi_1 \geq \pi_2 ... \geq 0, \sum_{i \in \Nats} \pi_i=1 \right\rbrace.
\)
For each bounded sequence $(\pi_i)_{i\in \Nats}$ of non-negative real numbers we denote by
$(\pi_i)^{\downarrow}_{i \in \Nats}$ the re-ordering of $(\pi_i)_{i\in \Nats}$ in a decreasing manner; we thus have that $(\pi_i)^{\downarrow}_{i \in \Nats} \in \mcal{S}$ if and only if $\sum_{i \in \Nats} \pi_i=1$. We now define a fragmentation operator on the space $\mcal{S}$, and then give the definition of a fragmentation process (FP).

\definition[Random Fragmentation Operator]{
Let $\text{Frag}(\cdot, \cdot)$ be a fragmentation operator defined as follows:
\vspace{-0.25cm}
\[
\text{Frag}(\vec{\pi}, (\vec{\bar{\pi}}^{(i)})_{i \in \mathbb{N}}) := \left(\pi_i \cdot \bar{\pi}^{(i)}_k\right)_{i,k \in \Nats}^{\downarrow}
\label{eq-def-frag}
\]
where $(\vec{\bar{\pi}}^{(i)})_{i \in \mathbb{N}}$ are \iid copies of some random sequence $\vec{\bar{\pi}}$. That is, for every integer $i$, $\mbox{Frag}(\cdot, \cdot)$ defines the distribution over the partitions of the $i$-th block $\pi_i$ of $\vec{\pi}$ induced by the $i$-th \iid copy $\vec{\bar{\pi}}^{(i)}$. The resulting partitions are the scaled sequences $\pi_i\cdot(\bar{\pi}^{(i)}_1,\bar{\pi}^{(i)}_2,\ldots)$. Collecting these partitions for each $\pi_i$ and rearranging them in decreasing order, we get the right hand side of Equation \eqref{eq-def-frag}.}

\definition[Random Fragmentation Process, FP]{We call an $\mcal{S}$-valued Markov process $\vec{\pi}(t)\defas (\pi_n(t))_{n\in\Nats}$, a (mass) \emph{fragmentation process} if the following two conditions hold:
\begin{enumerate}[i.]
\topsep=0pt \itemsep=0pt
\item $\vec{\pi}(0)=(1,0,0,\ldots)$.
\item For any $t, u\in \Reals^{+}$, conditioned on $\vec{\pi}(t)$, the random variable $\vec{\pi}(t+u)$ has the following distribution:
\vspace{-0.2cm}
\[
\vec{\pi}(t+u) \disteq \text{Frag}(\vec{\pi}(t), (\vec{\pi}^{(i)}(u))_{i \in \mathbb{N}})
\]
where $\disteq$ means equality in distribution.
\end{enumerate}
\label{def-fp}
}

In the fragmentation process, each sequence $\vec{\pi}(t)$ corresponds to a specific sorted split of a stick as brought in the stick-breaking example before. Intuitively, a fragmentation process can be understood through the stick-breaking example; in each splitting event the stick $\pi_i$ is replaced with a (possibly infinite) sequence of shorter sticks that sum to $\pi_i$. The splitting event is independent of the splitting time, which in a more general setting would be given by a deterministic function. We will assume all sticks split concurrently according to such a function.
The selection of the deterministic function used for the splitting rate, or the \emph{divergence function},
will be explained further in the following Section. Note for practical purposes, in this work we focus on the \emph{discrete} time FP, that is, splitting events are only allowed at discrete time steps (which corresponds to a fragmentation chain, c.f. \citet{Bertoin2006Random}).

\subsection{Nested Chinese Restaurant Processes}
\label{sec-crp}
The nested Chinese restaurant processes \citep[nCRP;][]{blei2010nested} is a Dirichlet ``path-reinforcing'' traverse of a tree where each data point starts at the root and descends to the leaves. More specifically, the first data point descends from the root and creates a new node with probability $1$; the same data point repeats this process up to a pre-defined depth resulting in a leaf node (obtaining a chain graph). A later data point $i$ starts from the root, and descends according to a Chinese restaurant processes (until maximum depth $L$). That is, if the data point reaches $\vec{\omega}$, it will either descend to an existing child or create a new child with probabilities:
\[
p(\vec{\omega}\omega_i&|\vec{\omega}) =\begin{cases}
n_{\vec{\omega}\omega_i}/(n_{\vec{\omega}\cdot}+\alpha(|\vec{\omega}|))
\text{~~~ descends to child $\vec{\omega}\omega_i$}\\
\alpha(|\vec{\omega}|)/(n_{\vec{\omega}\cdot}+\alpha(|\vec{\omega}|))
\text{ creates a new child}
\end{cases}
\label{eq-crp-prob}
\]
Here $n_{\vec{\omega}\omega_i}$ denotes the number of data points descending from node $\vec{\omega}$ to node $\vec{\omega}\omega_i$ for all data points preceding data point $i$, and $n_{\vec{\omega}\cdot}$ denotes a marginal count. This formulation leads to the well known ``rich get richer'' self-reinforcing property, which has been proved useful in various applications such as topic modelling and genetic mutation clustering \citep{teh2010dp}.

\textbf{Probability of the Combinatorial Structure}
For each node $\vec{\omega}$ we refer to the set of ancestor nodes $(\vec{\omega}'\text{: } \Delta \preceq\vec{\omega}'\preceq\vec{\omega})$ -- including the root and  $\vec{\omega}$ itself -- as a path. The probability of each path is simply the product of probabilities given in Equation \eqref{eq-crp-prob}:
\[
p(\Delta \to \vec{\omega})=\prod_{(\vec{\omega}'\omega_i\text{: } \vec{\omega}'\omega_i\preceq\vec{\omega})}p(\vec{\omega}'\omega_i|\vec{\omega}')
\label{eq-prob-path}
\]

For each node, we refer to the collection of child nodes  $\vec{\omega}\omega_i$, and the counts associated with each child node as its branching structure. Since the branching structure is created by a CRP, we can write down the probability of the combinatorial branching structure analytically
\[
g_{\vec{\omega}} &= p\big(n_{\vec{\omega}\omega_i}: ~\forall~ i~|~n_{\vec{\omega}\cdot}, ~\alpha\big) 
= \frac{\Gamma(\alpha)
\alpha^{K_{\vec{\omega}}}
\prod_{\vec{\omega}\omega_i} \Gamma(n_{\vec{\omega}\omega_i})}
{\Gamma(n_{\vec{\omega}\cdot}+\alpha)}
\label{eq-crp-joint}
\]
where $K_{\vec{\omega}}$ is the number of children nodes of $\vec{\omega}$, and $\alpha$ is the concentration parameter.  

\section{Dirichlet Fragmentation Processes}
\label{sec-dfp}
There exist many distributions satisfying the second condition set in Definition \ref{def-fp}, each leading to a distinct family of fragmentation processes with different properties. One notable example of such distributions is the Poisson--Dirichlet (PD) distribution and its 2 parameter extension\footnote{The 2-parameter PD distribution is also known as the Pitman--Yor process (PYP).} \citep{pitman1997two}.
The PD distribution and its extensions have been shown to be powerful Bayesian nonparametric tools for mixture models (e.g. the popular Dirichlet process (DP) mixtures). Motivated by this success of the PD distribution, in this paper we derive a \emph{Dirichlet fragmentation process} (DFP) defined as follows.

\definition[DFP]{We call a fragmentation process a \emph{Dirichlet fragmentation process} if at each time $t$ the Frag operator induces a Poisson-Dirichlet distribution over the partitions.
\label{def:dfp}
}

A useful property of the random fragmentation process is that it satisfies the Markov property -- given a stick $\vec{\omega}$, subsequent fragmentation events are independent from $\vec{\omega}$'s ancestors in the tree.

\subsection{Recursive Stick-Breaking Construction}
\label{sec:sb}
We gave an imprecise description of the stick breaking process in Section \ref{sec-frag-intro}; now we give a formal definition to the process and use it as a constructive procedure for sampling from the Dirichlet fragmentation process. The stick-breaking process defined by \citet{sethuraman1994a}
is a constructive way for drawing samples from the DP. A random probability measure $G$ can be drawn from a DP given a base probability measure $H$ and concentration parameter $\alpha$ using a sequence of beta draws:
\[
\begin{split}
G=\sum_{k=1}^{\infty}\pi_k\delta_{\phi_k},\qquad
\pi_k =\nu_k\prod_{i=1}^{k-1}(1-\nu_{i}), \qquad\\
\nu_k \distiid \BetaDist(1,\alpha),\qquad
 \phi_k \dist H.\qquad\qquad
\end{split}
\]
This can be viewed as taking a stick of unit length and breaking it at a random location. We call the left side of the stick $\pi_1$ and then break the right side at a new place, call the left side of this new break $\pi_2$. We then continue this process of ``keep the left piece and break the right piece again''. \citet{sethuraman1994a} showed that the sequence of weights obtained from the stick breaking process $(\pi_1,\pi_2,\ldots)$ distributes according to the  Poisson-Dirichlet distribution \citep{pitman1997two}. Thus the stick breaking procedure can be used as a Dirichlet Frag operator. This a is a useful property since we can apply this stick breaking Frag operator in a recursive way to induce a tree structure. This property has been noted and studied by \cite{adams2010tree}. Here we provide a modified tree structured stick breaking procedure and use it as a way for sampling from the DFP.

Now we describe the recursive stick breaking process. The first step is to sample a beta random variable $\nu_{\vec{\omega}} \dist \BetaDist(1, \alpha)$ for each node in the tree with the exception of the root node. Then the length of the stick associated with node $\vec{\omega}\omega_i$ is given by
\begin{gather}\label{eq:4.2}
\pi_{\vec{\omega}\omega_i}=\pi_{\vec{\omega}}\nu_{\vec{\omega}\omega_i}
\prod_{k=1}^{\omega_i -1}(1-\nu_{\vec{\omega}k}),
\end{gather}
where $\pi_{\vec{\omega}}$ is the stick length of the parent node.
Through multiplying over beta variables of all prefixes of $\vec{\omega}$, the recursive definition given in Equation \eqref{eq:4.2} can be unpacked as
\begin{gather}
\pi_{\vec{\omega}}=\prod_{\vec{\omega}'\omega_i\preceq\vec{\omega}}
\nu_{\vec{\omega}'\omega_i}\prod_{k=1}^{\omega_i-1}(1-\nu_{\vec{\omega}'k}).
\end{gather}
More generally, the concentration parameter $\alpha$ is allowed to vary for different nodes. For example, $\alpha$ can be a function of the depth of a given node, denoted by $\alpha(|\vec{\omega}|)$. When the concentration parameter is infinitesimal for each node (\eg $\alpha(|\vec{\omega}|)=a(t_{\omega})dt$, whereas $t$ can be seen as a fictitious time associated node $\omega$), and the maximum depth of tree is sufficiently large, the  recursive stick breaking will generate binary trees with probability $1$. This special case of the DFP is known as the Dirichlet diffusion tree \citet[DDT, ][]{neal2003density}.
Following a convention first introduced by \citet{neal2003density}, we shall call this function $\alpha(|\vec{\omega}|)$ the divergence function\footnote{An example of such a function is:
\[
\label{eq-div-fun}
\alpha(l)&=a\big((l+1)/L\big)-a\big(l/L\big),
\]
where $L$ is the number of discrete time steps, and $a$ is defined as:
\(
a(s)=\int_0^{s}c/(1-s)ds,
\label{eq-div-fun-alpha}
\)
for some hyper parameter $c$.}. The recursive stick-breaking process and the tree node indexing scheme are illustrated in Figure \ref{fig-re-stick-breaking}.

\begin{figure*}[t]
\vspace{-0mm}
\centering
\includegraphics[trim = 15mm 10mm 15mm 10mm, clip, width=0.45\textwidth]{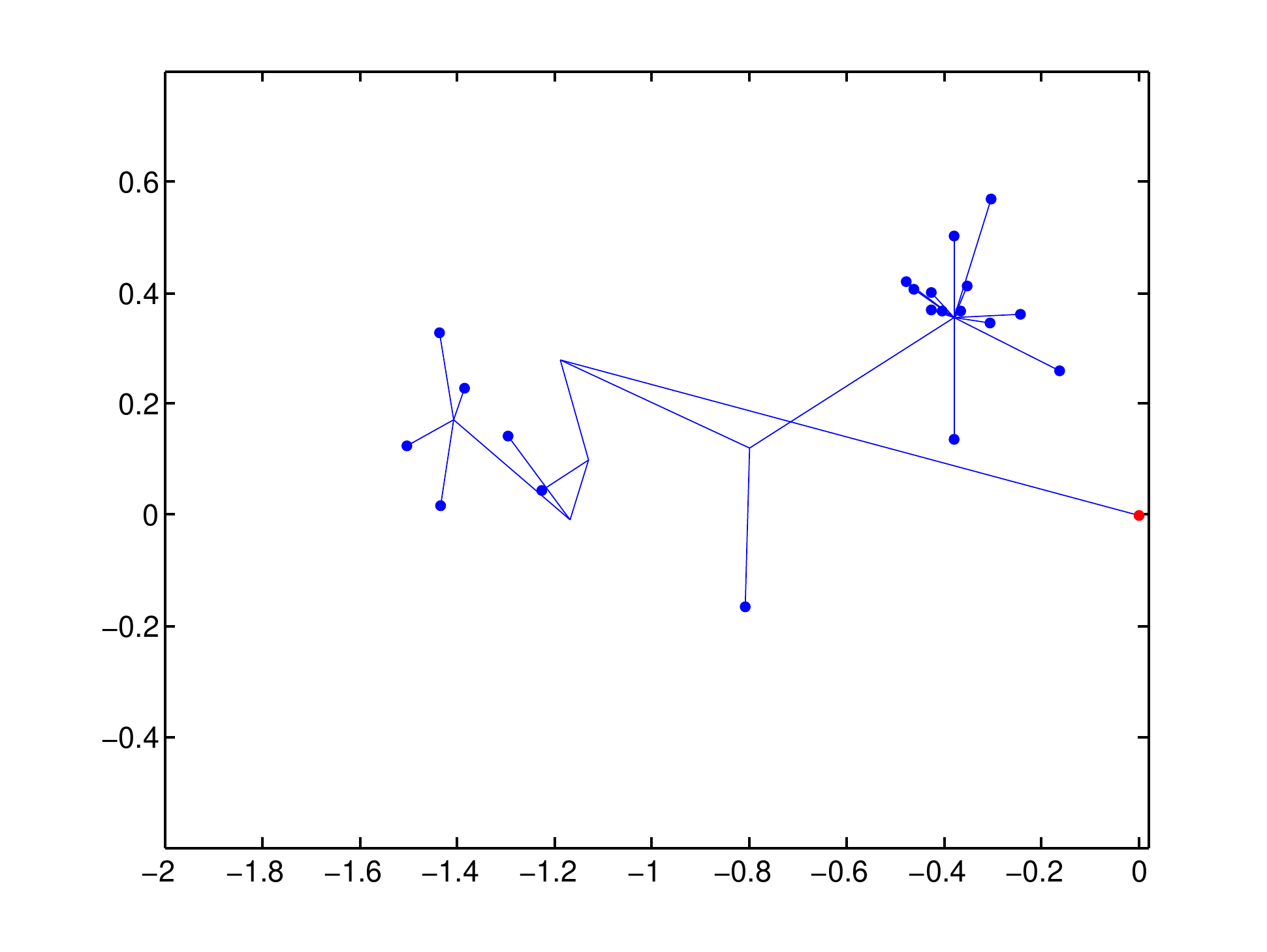}
\includegraphics[trim = 15mm 10mm 15mm 10mm, clip, width=0.45\textwidth]{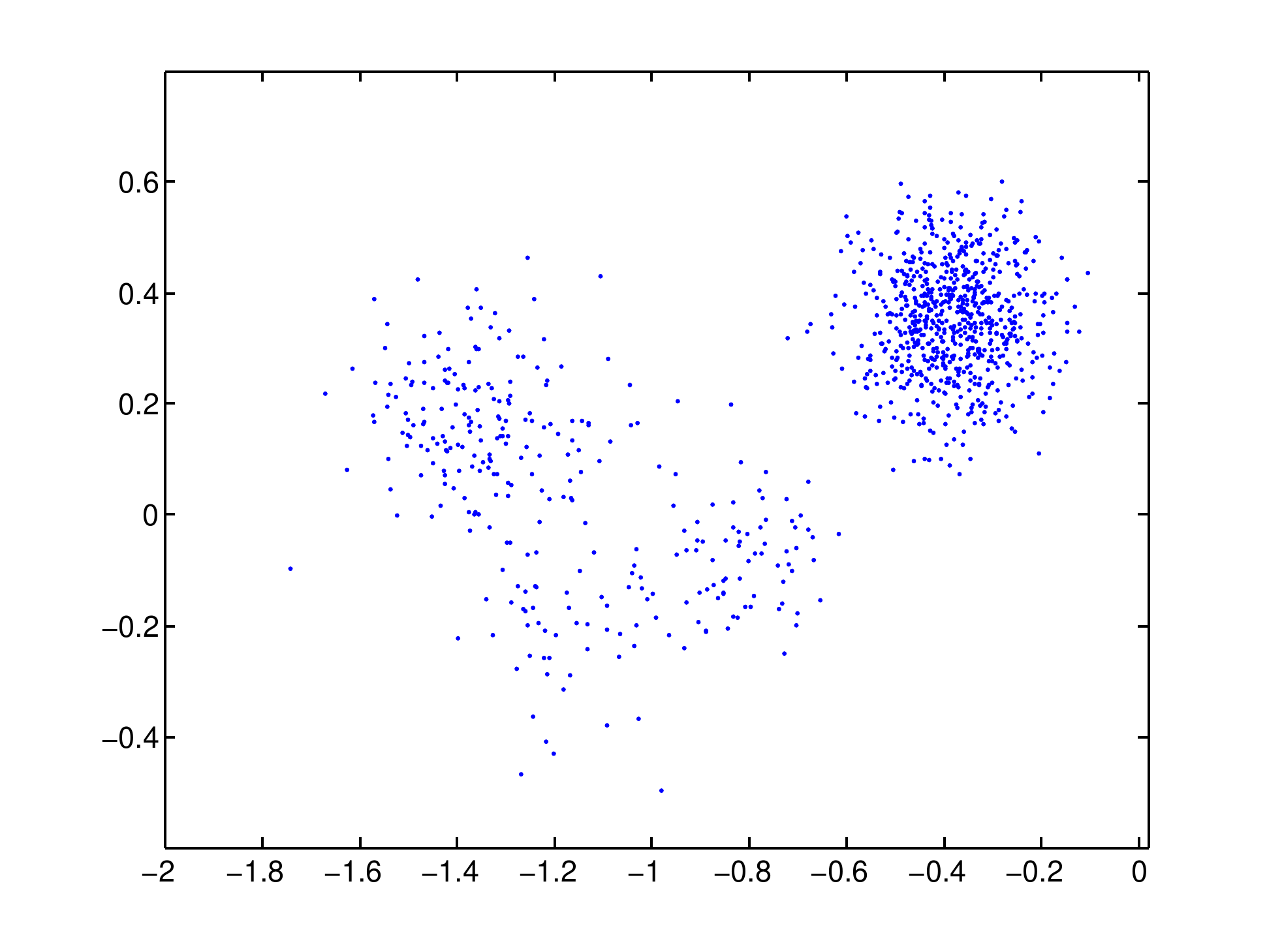}
\vspace{-0.2cm}
\caption{DFP Gaussian Diffusion Examples. Generation of a two-dimensional dataset from the Gaussian diffusion with the number of discrete time steps $L=40$, $\sigma=1$ and $\alpha(l)$ given by Equation \eqref{eq-div-fun} in the footnote. The plot on the left shows the first 20 data points generated, along with the underlying tree structure. The right plot shows 1000 data points obtained by continuing the procedure beyond these 20 data points.}
\label{fig-gaussian-difussion-ex}
\end{figure*}

\subsection{Parent-Child Transition Operators}
\label{sec-transition-op}
Recall that for the Dirichlet process mixture model an unbounded number of partitions is generated where each partition is labelled with some parameter $\phi_k \dist H$. Given the generated data partition and corresponding labels, each data point is assumed to arise as a draw from a distribution $F(y|\phi_k)$, where $\phi_k$ is the $k$'th component label from which $y$ is generated. In the DFP we continue to assume that the data are generated independently given the latent labelling, but take advantage of the tree-structured partitioning of the data. That is, the distribution over the parameter at node $\vec{\omega}\omega_i$, denoted $\phi_{\vec{\omega}\omega_i}$, should depend on its parent $\vec{\omega}$. This parent-child dependence will be captured through a \emph{Transition Operator}, denoted $T(\phi_{\vec{\omega}\omega_i} \gets \phi_{\vec{\omega}})\defas p(\phi_{\vec{\omega}\omega_i} | \phi_{\vec{\omega}})$.
For example, the Gaussian transition operator is given by
\[
T(\phi_{\vec{\omega}\omega_i} \gets \phi_{\vec{\omega}})
=\mathcal{N}(\phi_{\vec{\omega}}, \sigma^2),\;\;\;
p(\Delta) = \mathcal{N}(0, \sigma^2)
\label{eq-trans}
\]
where $p(\Delta)$ denotes the parameter distribution of the root node. An example of $1000$ data points sampled from the DFP with a Gaussian transition operator is given in Figure \ref{fig-gaussian-difussion-ex}.

\section{A DFP Mixture Model}
\label{sec-model}
Given a DFP prior over the tree structure, we can obtain a hierarchical infinite mixture model by coupling the model with a mixture model component likelihood function, for example a Gaussian data distribution \ie
\[
F(y|\phi_{\vec{\omega}})=\mathcal{N}(\phi_{\vec{\omega}}, \sigma^2).
\]
Here the subscript $\vec{\omega}$ denote the index of the leaf node associated with data point $y$.
We assume the dimensionality of the data to be $1$ to keep notation simple. We will use this notation in the remaining part of the paper since the extension to arbitrary dimensionality is straightforward.

\begin{figure*}[tt]
  \centering
  \includegraphics[trim = 10mm 5mm 15mm 10mm, clip, width=0.22\linewidth]{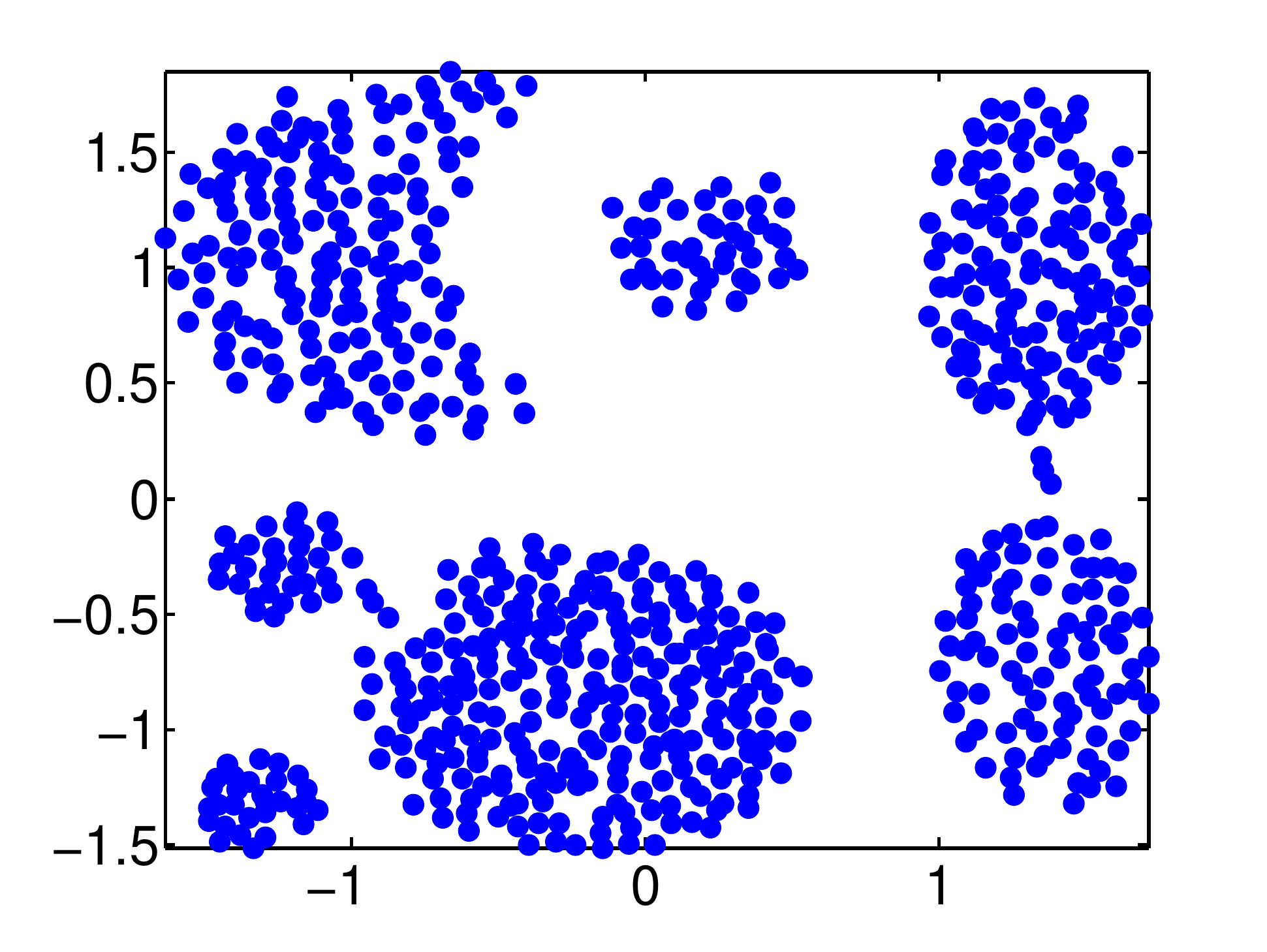}
  \includegraphics[trim = 21mm 20mm 15mm 20mm, clip, width=0.55\linewidth, height=25mm]{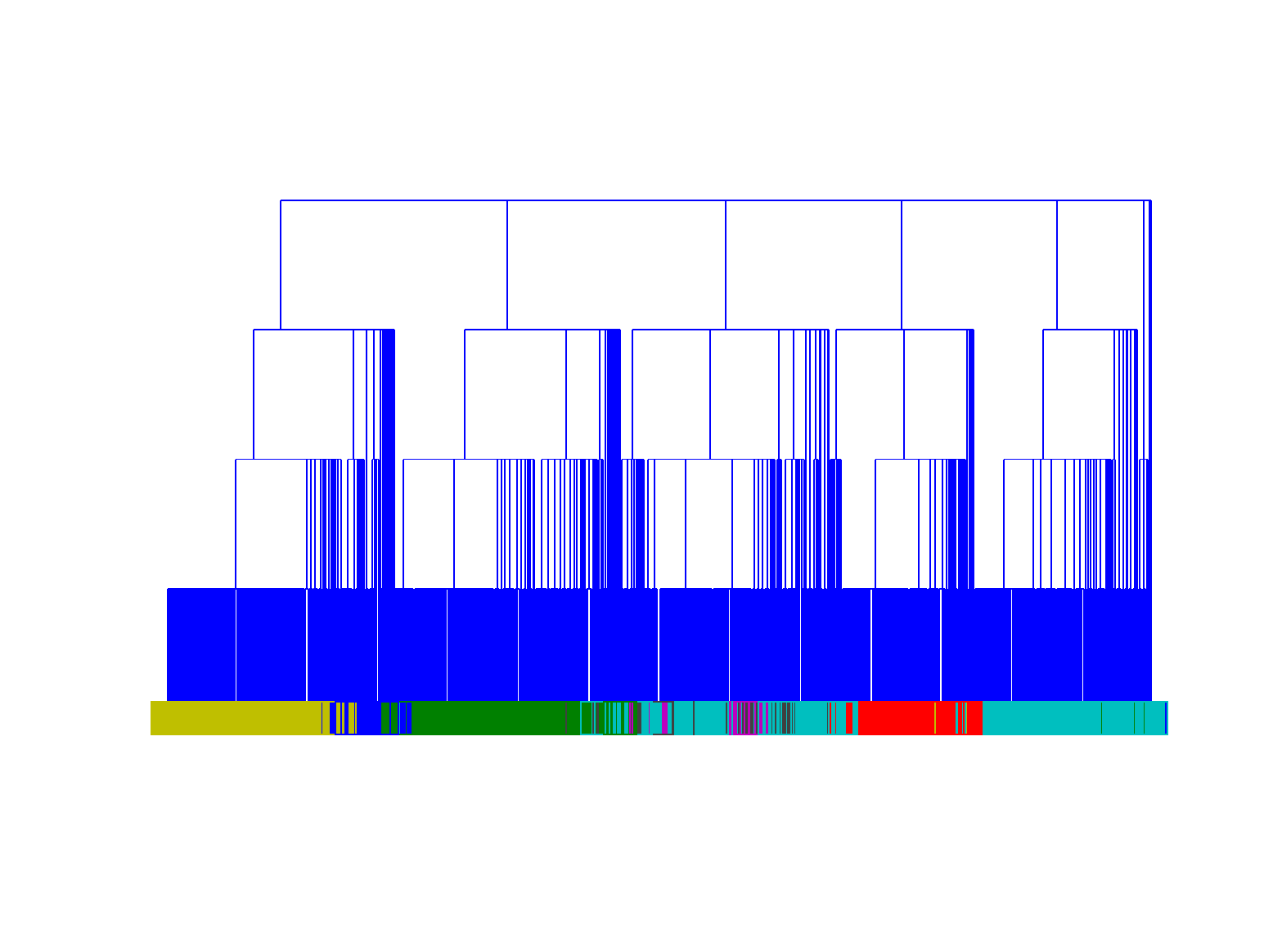}
  \includegraphics[trim = 20mm 10mm 15mm 10mm, clip, width=0.21\linewidth]{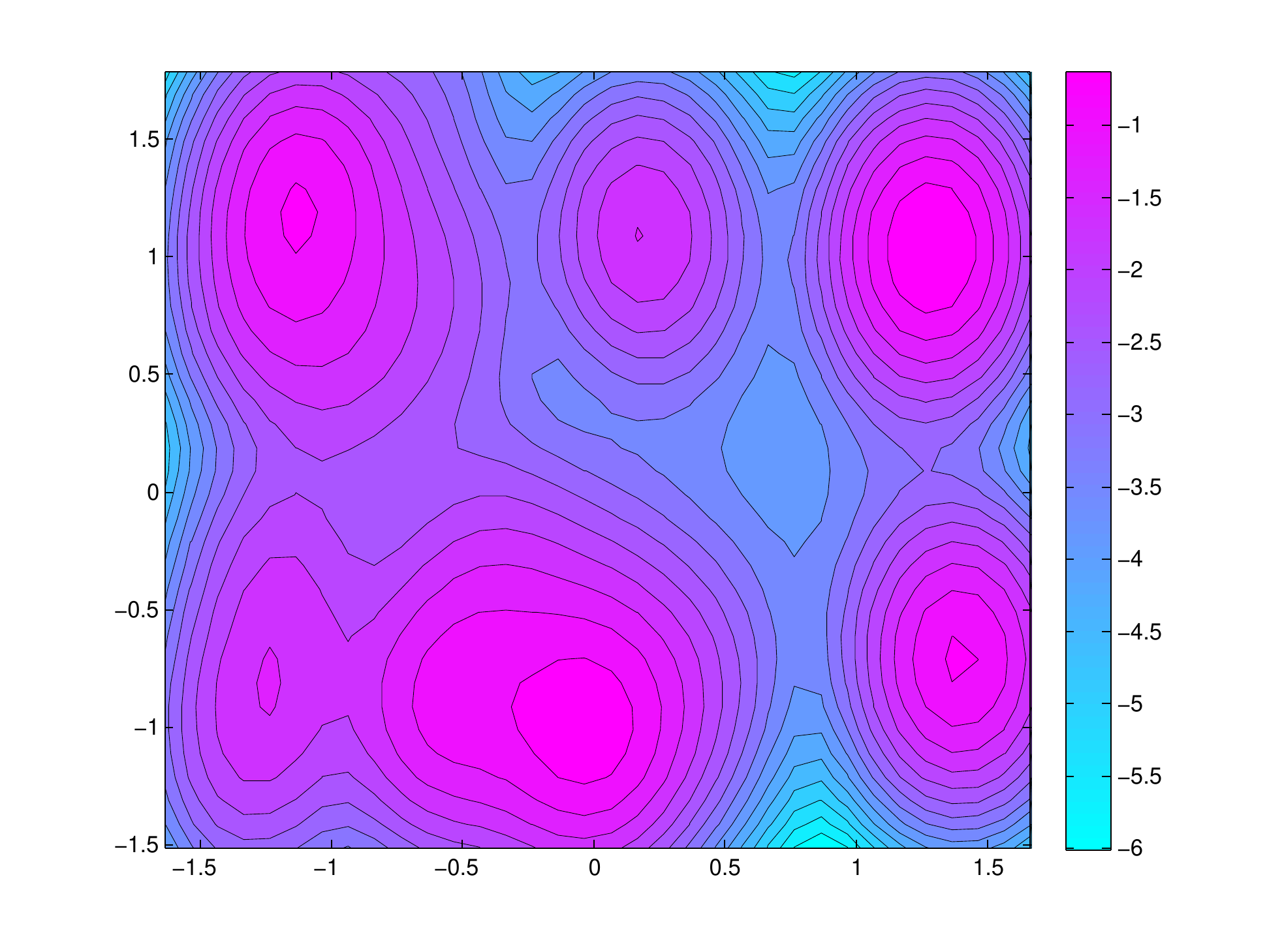}\\
  \vspace{2mm}
  \includegraphics[trim = 10mm 5mm 15mm 10mm, clip, width=0.22\linewidth]{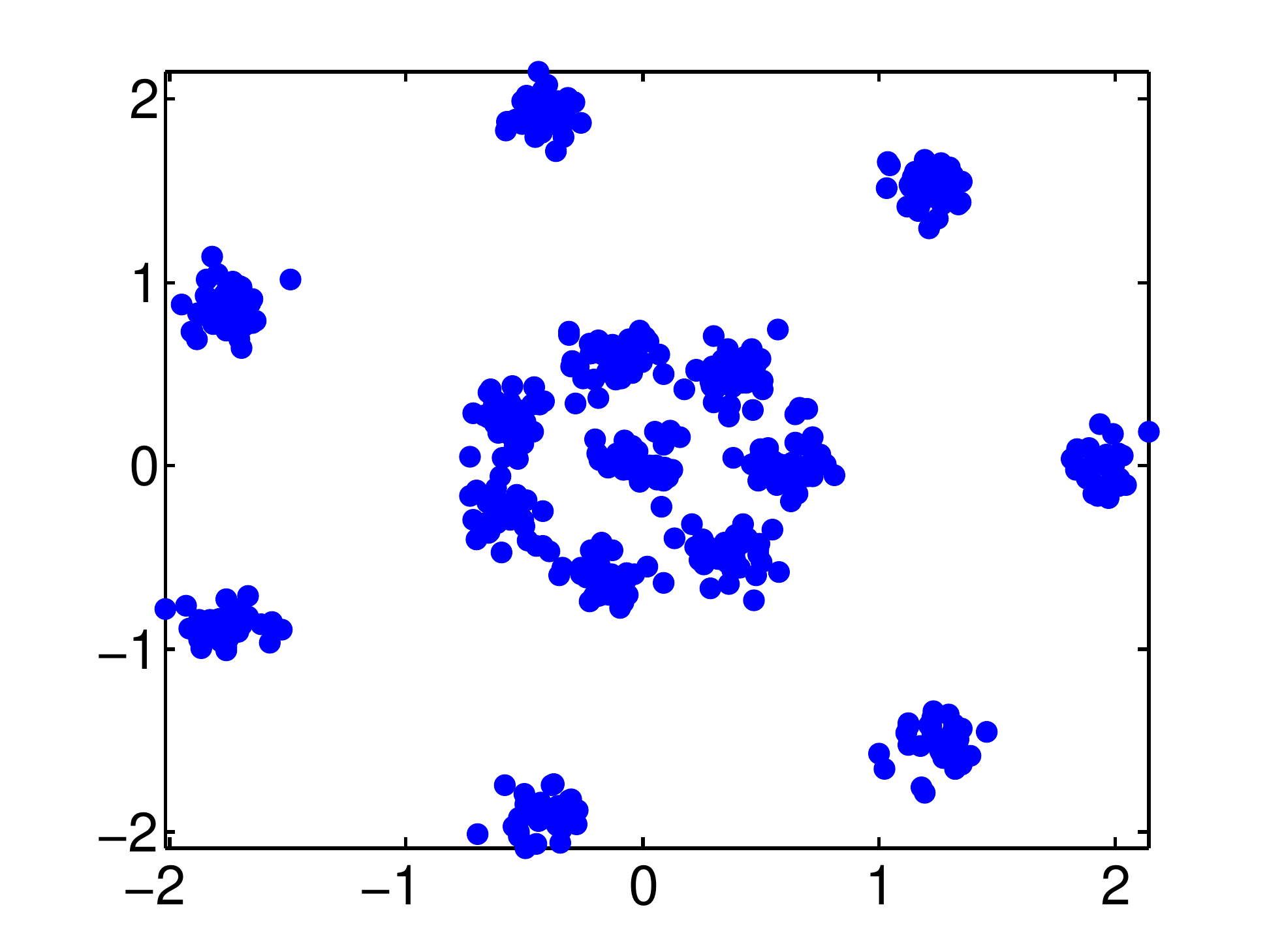}
  \includegraphics[trim = 20mm 20mm 15mm 20mm, clip, width=0.55\linewidth, height=25mm]{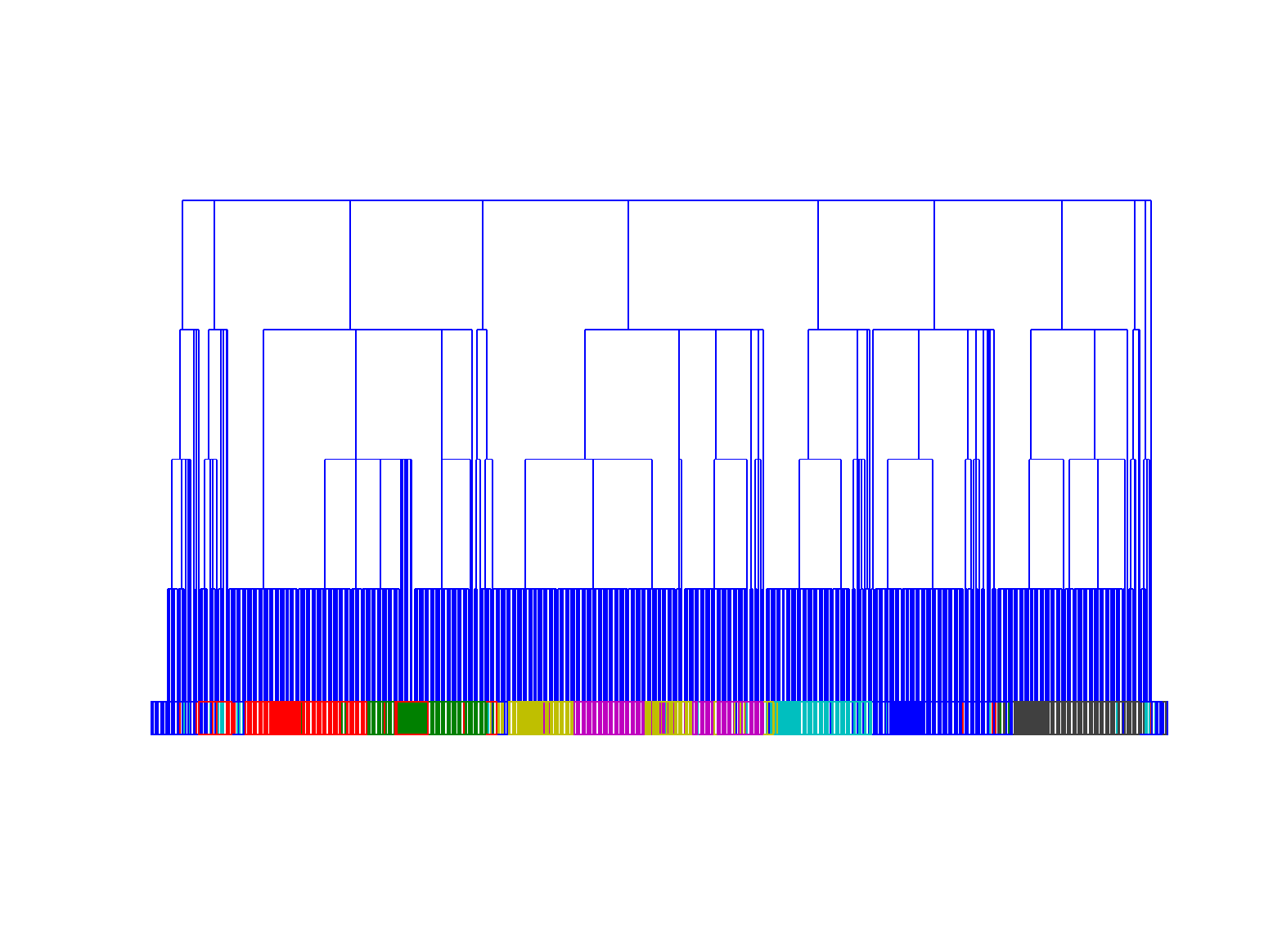}
  \includegraphics[trim = 20mm 10mm 15mm 10mm, clip, width=0.21\linewidth]{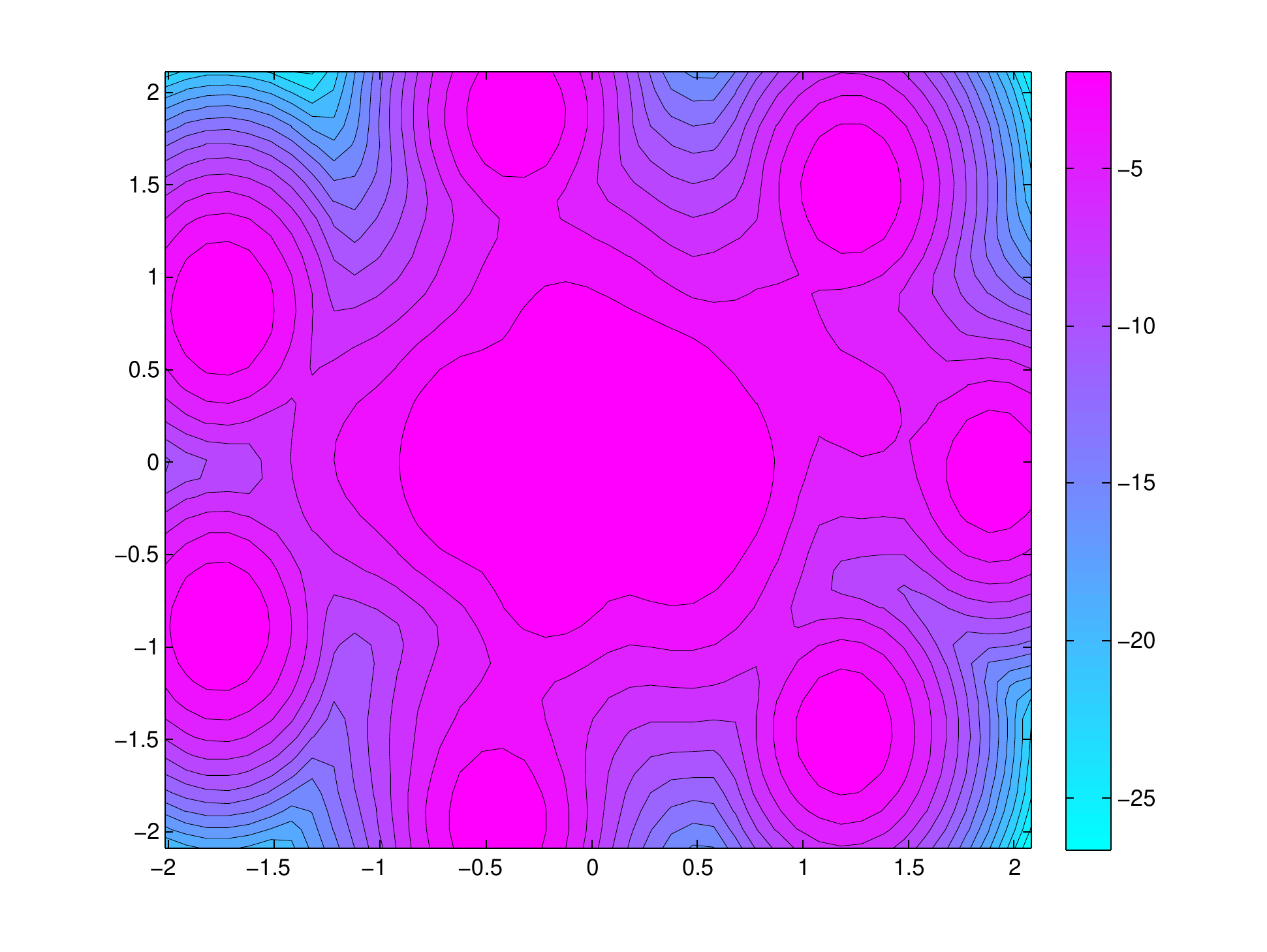}
      \vspace{-0.2cm}
\caption{Results on Aggregation (top row), and R15 datasets (bottom row). The left plot shows the original data; the middle plot gives trees sampled from the posterior conditioning on the training data; the right plot shows the predictive densities resulting from our DFP mixture model.}
\label{fig-further-results}
\end{figure*}

\section{Inference by Gibbs Sampling}
\label{sec-inference}
Recall our variables of interest; the variables $y_i$ are our observations, and we let $z_i$ denote the node (i.e. mixture component) from which $y_i$ was generated -- each $y_i$ is assumed to arise as a draw of $F(y_i|\phi_{z_i})$.
Here the vector $\vec{\phi}=(\phi_{\vec{\omega}})$ stores the parameters of each node. We use $n_{\vec{\omega}\cdot}$ to denote the number of leaves descended from node $\vec{\omega}$, and $K_{\vec{\omega}}$ to denote the number of children of node $\vec{\omega}$. Furthermore, we use $\vec{\omega}k$ to denote the $k$'th child of $\vec{\omega}$.

Let $\vec{y}=y_{1:N}$ be the sequence of data items, $\vec{y}_{\vec{\omega}}=(y_i\text{: } z_i=\vec{\omega})$ be the sequence of data items generated from node $\vec{\omega}$, and $\mathbf{z}=z_{1:N}$ be the sequence of nodes generating $\vec{y}$. We attach a superscript to a set of variables or a count (e.g. $\mathbf{y}^{-i}$ or $n_{\vec{\omega}}^{-i}$) to denote the removal of the variable corresponding to the superscripted index from the variable set or from the calculation of the count. In our examples $\mathbf{y}^{-i}:=\mathbf{y}\backslash y_i$ and $n_{\vec{\omega}}^{-i}$ is the number of observations (i.e. leaves) ultimately reached by node $\vec{\omega}$, leaving out data point $y_i$.

In the case of the Gaussian observation model, which is conjugate to the distribution of the leaf parameters, we integrate out the leaf and internal parameters $\phi$ in the sampling schemes. Denote the conditional density of $y_{i}$ under leaf node $z$ given all data points except $y_i$ as $f_z^{-y_i}(y_i)$.
The non-conjugate case can be tackled by adapting similar techniques to the ones developed for non-conjugate DP mixtures \citep{neal2000markov}.

Finally we specify priors on the hyper-parameters of the divergence function (Equation \eqref{eq-div-fun} in the footnote), $c$, and the diffusion precision $\tau$ (the inverse of $\sigma^2$ in Equation \eqref{eq-trans}):
\begin{align}
c \dist \GammaDist(a_c,b_c), \;\;\;
\tau \dist \GammaDist(a_{\tau},b_{\tau})
\end{align}
Here $\GammaDist(a,b)$ is a Gamma distribution with shape $a$
and rate $b$. In all experiments we used $a_c=1,b_c=1,a_{\tau}=1,b_{\tau}=1$. Next we describe a Gibbs sampler for the DFP.

\textbf{Step 1: Sampling $\mathbf{z}$}. This can be realised by
\[
\begin{split}
&p(z_i=\vec{\omega}|\mathbf{z}^{-i}, c, \tau) \\
&\propto
\begin{cases}
p_{\vec{\omega}}^{-i}f_{\vec{\omega}}^{-y_i}(y_i) &\text{if $\vec{\omega}$ is an existing leaf node}\\[0.5em]
p^{-i}_{\vec{\omega}'}\frac{\alpha_{\vec{\omega}'}}{\alpha_{\vec{\omega}'}+n_{\vec{\omega}'\cdot}} f_{{\vec{\omega}'}}^{-y_i}(y_i) &\text{if $\vec{\omega}$ is a new leaf node}
\end{cases}
\end{split}
\]
with $\vec{\omega}'$ parent of $\vec{\omega}$.

Here $p_{\vec{\omega}}^{-i}$ is the probability of reaching node $\vec{\omega}$ from the root node leaving out $y_i$ (as defined in Equation \eqref{eq-prob-path}), and $\alpha_{\vec{\omega}}$ is the divergence function evaluated at depth $|\vec{\omega}|$. Intuitively, the above equation defines the two ways that $y_i$ can be generated. In the first way, data item $i$ follows an existing branch until it reaches a leaf node $\vec{\omega}$, which has probability $p^{-i}_{\vec{\omega}}$. Then this probability is multiplied with the likelihood term, giving us the total probability that $y_i$ is generated from node $\vec{\omega}$. In the second way, data item $i$ initially follows an existing branch until it reaches (internal) node $\vec{\omega}'$, then it diverges from the current branch and creates a new leaf node, for which the total probability is simply the product of the probability of reaching node $\vec{\omega}'$, and the probability of diverging from $\vec{\omega}'$. Lastly, multiplied with a likelihood term, this gives us the probability of $y_i$ being generated from a new node. Note that updating the leaf assignment of each data point $y_i$ will also update the count vector ${n}_{\vec{\omega}\cdot}$, and vice versa. In fact, this is the only way that $\mathbf{z}$ influences the other variables, i.e. $\vec{\phi}$ and $c$.

\textbf{Step 2: Sampling divergence function hyperparameter $\textbf{c}$}. The probability of the tree structure given the divergence function is simply the product of the probabilities of the branching structures for each internal node. Since at each internal node $\vec{\omega}$ the process of descending to the children follows a CRP, the probability of a branching structure for each internal node $g_{\vec{\omega}}$ is given by Equation \eqref{eq-crp-joint}. Coupled with the gamma prior, the Gibbs conditional probability for $c$ is
\[
p(c|\tau, \mathbf{g}, \mathbf{n}, \vec{\phi})
\propto \GammaDist\left(a_c,b_c\right)
\prod_{(\vec{\omega} \text{: all internal nodes})}g_{\vec{\omega}}.
\]

\textbf{Step 3: Sampling the precision {\boldmath$\tau$}}. It is straightforward to sample $\tau$ given the node parameters $\vec{\phi}$. The probability of all node parameters $p(\vec{\phi})$ is simply the product of a set of Gaussians, since each node's parameter distribution $p(\phi_{\vec{\omega}{\omega_i}}|\phi_{\vec{\omega}})$ is Gaussian. Coupled with a gamma prior, the Gibbs conditional probability for the precision $\tau$ is given by
\[
p(&\tau|c, \mathbf{g}, \mathbf{n}, \vec{\phi})
\propto \GammaDist(a_{\tau},b_{\tau}) \times \nonumber\\
&\prod_{\scalebox{0.6}{$(\vec{\omega}\text{: all internal nodes})$}}
\prod_{\scalebox{0.6}{$(\vec{\omega}\omega_i\text{: children of }\vec{\omega})$}}\GammaDist\left(1,\frac{(\phi_{{\vec{\omega}\omega_i}}-\phi_{\vec{\omega}})^2}{2}\right).
\]

In summary, for each observation the proposed Gibbs sampler iteratively samples a path leading to it conditioned on paths leading to remaining observations (note this is different from the Gibbs sampler for the nCRP topic model, which samples path leading to each observation in two separate steps, see \citet{blei2010nested} for details). Most existing inference procedures for trees employ a ``prune-graft'' algorithm; that is, first remove a subtree and then propose to re-attache the sub-tree elsewhere. The proposal is then accepted or rejected using an MH step. As we will show in the following section, empirically  this Gibbs sampler results in significantly improved performance when compared to state-of-the-art models using this ``prune-graft'' inference for both hierarchical clustering and density estimation.

\begin{figure}[tt]
  \centering
  \begin{minipage}{0.7\textwidth}
  \includegraphics[trim = 5mm 5mm 5mm 0mm, clip, width=0.45\textwidth]{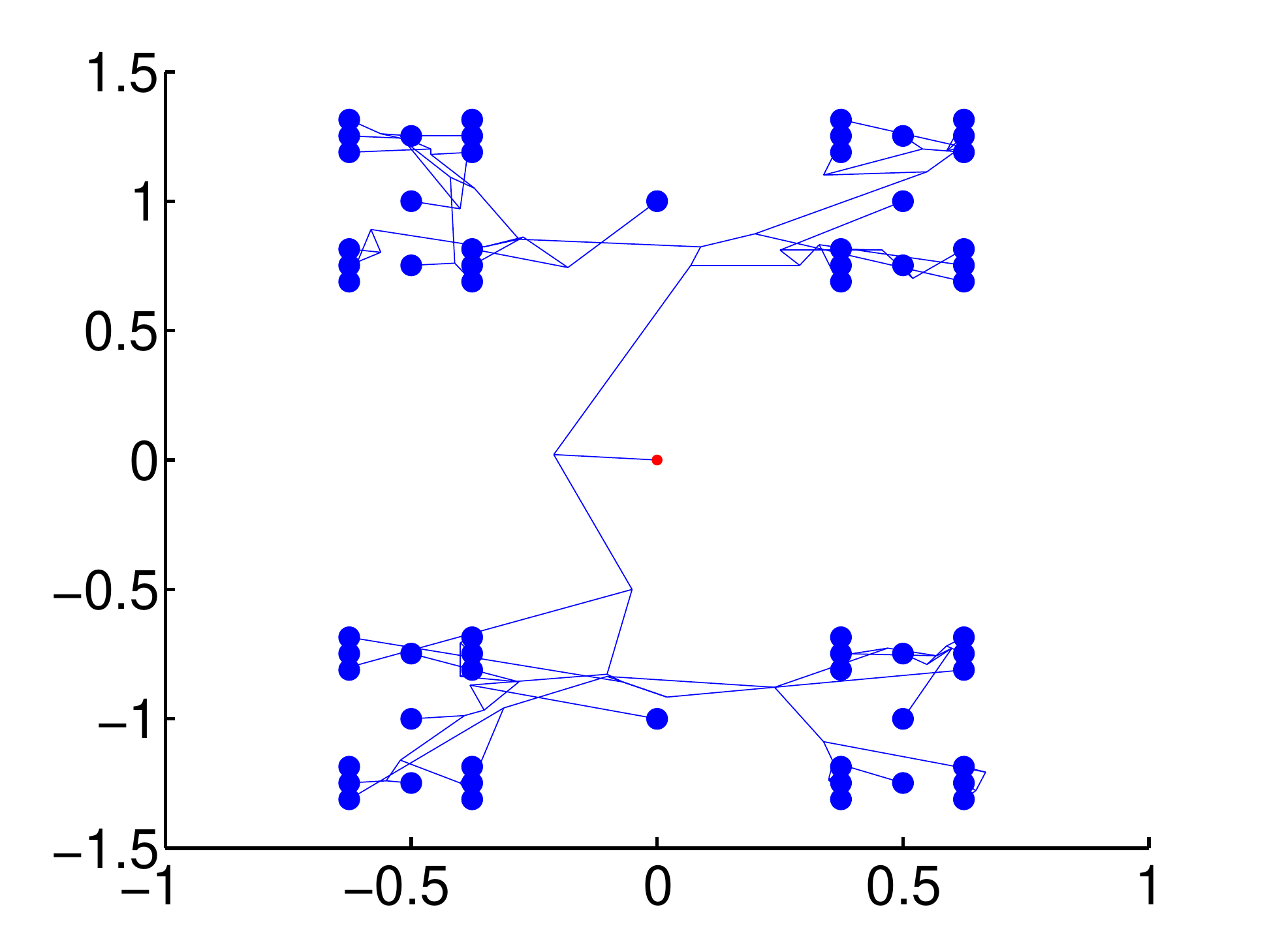}
  \includegraphics[trim = 5mm 5mm 5mm 0mm, clip, width=0.45\textwidth]{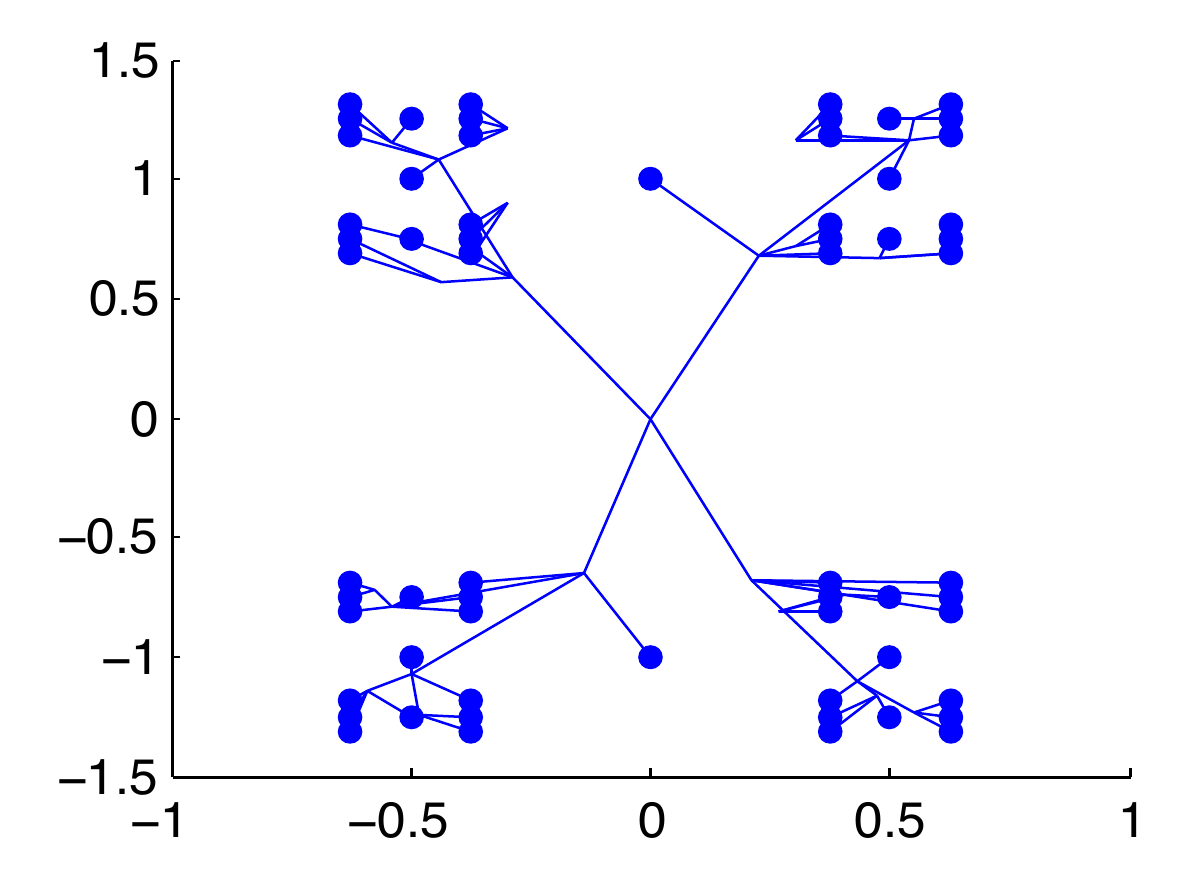}
  \end{minipage}
  \begin{minipage}{1\textwidth}
  \caption{Hierarchical clustering results on the Synthetic dataset. Left:  tree structures sampled from the DDT model conditioning on the data. Right:  tree structures sampled from the DFP mixture model conditioning on the data. }
\label{fig:dfp:ddt}
  \end{minipage}
  ~\vspace{-0.3cm}
\end{figure}

\section{Experiments}
\label{sec-results}
In this section we describe two sets of experiments to highlight the two aspects of the discrete time DFP mixture model: its hierarchical nature and its nonparametric density modelling nature. To demonstrate the hierarchical nature of the DFP we compare the model to the agglomerative clustering algorithm. For the DFP, we implemented the inference algorithm described in Section \ref{sec-inference}.  The software implements the discrete DFP with arbitrary depth, and is available at [URL]. We use Neal's Flexible Bayesian Modelling (FBM) package for the DDT and Matlab's implementation for the agglomerative clustering algorithm.

\begin{table*}[tt]
\centering
\begin{tabular}{lllll}
\toprule
DATASET    & GMM & DPM & DDT & DFP mixture \\
\midrule
R15  & $-2.127\pm0.158$    & $-0.759\pm0.122$ &  $-0.861\pm0.123$  & $\mathbf{-0.705\pm0.086}$      \\
D31  & $-2.593\pm0.036$    &  $-1.790\pm0.040$  &  $-1.798\pm0.044$ & $\mathbf{-1.654\pm0.022}$      \\
AGGR.  &  $-2.151\pm0.076$    & $-2.064\pm0.063$ &  $-2.091\pm0.057$  & $\mathbf{-1.431\pm0.008}$      \\
MACA.   &$-15.039\pm0.584$  &$-15.145\pm0.611$&$-14.816\pm0.546$&$\mathbf{-12.725\pm0.127}$\\
CCLE   &$-4.8183\pm0.947$  & $-4.6036\pm0.331$ &$-3.7266\pm0.457$&$\mathbf{-2.825\pm0.495}$\\
\bottomrule
\end{tabular}
\caption{Predictive log likelihood ($\log_e$) for GMM, DP mixture, DDT, and DFP mixture.}
\vspace{-0.3cm}
\label{tab-results}
\end{table*}

\subsection{Hierarchical Clustering}

First we compare the DFP mixture model to the agglomerative clustering algorithm.
We performed experiments on four datasets (one hand crafted synthetic dataset and three real datasets).
The real datasets we used are R15 (600 examples, 2 attributes \citet{veenman2002maximum}), Aggregation (referred to as AGGR, 788 examples, 2 attributes, \citet{veenman2002maximum}), and Glass (214 examples, 7 classes, 9 attributes).
For the synthetic dataset, trees sampled from the posterior of the DFP mixture model and the DDT conditioning on the training data are shown in Figure \ref{fig:dfp:ddt}. Both methods find a good hierarchical clustering of the data items.
While the DDT is forced to choose a binary branching structure over the clusters, the DFP can represent a more parsimonious solution. Such parsimonious solutions are more interpretable and potentially lead to better explanations for the data. Similar results are also observed for the real datasets. The results on the AGGR and R15 datasets are shown in Figure \ref{fig-further-results}. As we can see from Figure \ref{fig-further-results}, most data points with the same class label are merged in the first level of the DFP mixture model, which leads to a clean summary of the structure of the data.

Furthermore, in order to assess the quality of these hierarchical clustering results, we also computed the tree purity score for various algorithms on the Glass dataset; the tree purity score was introduced by \citet{heller2005bayesian} and motivated as a reasonable metric for evaluating hierarchical clustering algorithms. On the Glass dataset the purity scores are 0.5064 (DFP), 0.4815 (agglomerative, average linkage), and 0.4568 (DDT). The result of the agglomerative algorithm are consistent with those reported in \citet{heller2005bayesian}.
However, while \citet{heller2005bayesian}'s Bayesian Hierarchical Clustering algorithm exhibits lower purity score when compared to the agglomerative algorithm on the Glass dataset, the DFP mixture model produces a slightly better one.

\subsection{Density Estimation}
\label{densitymodelling}

To evaluate the power of the DFP in density estimation, we compare the DFP mixture model to traditional mixture models including the Gaussian Mixture Model (GMM), the Dirichlet Process Mixtures (DPM), and the Dirichlet Diffusion Tree (DDT) over 5 datasets. The 5 datasets we used are the macaque skull measurements (MACA, 228 examples, 10 attributes), R15, Aggregation, D31 (3100 examples, 2 attributes), and the Cancer cell line encyclopedia (CCLE, 504 examples, 24 attributes). In particular, the CCLE dataset consists of measurements of the sensitivity of 504 cancer derived cell lines to 24 drugs.
Such data has the potential to help biologists understand the relationship between different cancer types and drug effects, and to aid in clinical practice \citep{barretina2012cancer}.

For all datasets we train each model using $90\%$ of the data and report the predictive log likelihood for the remaining $10\%$ of the data. For the DFP, we set the depth of the tree at $L=4$. For all methods under comparison, we run the MCMC inference algorithm until the predictive log likelihood for the train data converges. 
As shown in Table \ref{tab-results}, on all datasets the DFP mixture model obtains the highest predictive log likelihood. For the MACA dataset, the DFP mixture model outperforms all previous models: the performance of the model is $2.5$ orders better (on $\log_e$ scale). This is a significant improvement as previous attempts on the same dataset only obtained a small improvement, as reported in \citet{knowles2011pitman} and \citet{murray2008gaussian}. 
The improvement of the DFP over existing methods is consistent with all other datasets we tried, in particular, the performance on the CCLE is about $1$ order better.

 \section{Discussion}\label{sec:related_work}
This paper have presented the Dirichlet fragmentation process for modelling tree structures. The DFP is derived as a useful variant of fragmentation processes, and is connected to a number of existing models such as \citet[DDT][]{neal2003density}, \citet[nCRP][]{blei2010nested}, \citet[TSSB][]{adams2010tree}, \citet[PYDT][]{knowles2011pitman}, \citet[nDP][]{rodriguez2008nested}. Particularly, we derived a simple hierarchical mixture model based on the DFP, and an efficient Gibbs-style sampler.  This DFP hierarchical mixture model generalises the popular Dirichlet process mixture model. Unlike the latter, which partitions data into a flat layer of clusters, the DFP mixture model organises clusters into a tree structure. Not only this provides more interpretable summary of the data, but also leads to significantly better accuracy as demonstrated in the density estimation experiments. 
Future theoretical work will study the connection between the DFP and hierarchical DPs, and extends the DFP to model group data and sequential data.

\begin{spacing}{0.9}
\bibliographystyle{unsrtnat}
\bibliography{dfp-ee} \end{spacing}
\end{document}